\documentclass[sn-mathphys]{sn-jnl}

\jyear{2022}%

\theoremstyle{thmstyleone}%
\theoremstyle{thmstyletwo}%

\theoremstyle{thmstylethree}%

\begin{document}

\title[Deep Multi-Agent Reinforcement Learning with Hybrid Action Spaces based on Maximum Entropy]{Deep Multi-Agent Reinforcement Learning with Hybrid Action Spaces based on Maximum Entropy}


\author*[1]{\fnm{Hongzhi} \sur{Hua}}\email{hongzhihua@cqu.edu.cn}

\author[2]{\fnm{Kaigui} \sur{Wu}}\email{kaiguiwu@cqu.edu.cn}
\equalcont{These authors contributed equally to this work.}

\author[3]{\fnm{Guixuan} \sur{Wen}}\email{guixuanwen@cqu.edu.cn}
\equalcont{These authors contributed equally to this work.}

\affil*[1]{\orgdiv{College of Computer Science}, \orgname{Chongqing University}, \orgaddress{\street{Shazheng Street}, \city{Chongqing}, \postcode{400044}, \country{China}}}

\affil[2]{\orgdiv{College of Computer Science}, \orgname{Chongqing University}, \orgaddress{\street{Shazheng Street}, \city{Chongqing}, \postcode{400044}, \country{China}}}

\affil[3]{\orgdiv{College of Computer Science}, \orgname{Chongqing University}, \orgaddress{\street{Shazheng Street}, \city{Chongqing}, \postcode{400044}, \country{China}}}


\abstract{Multi-agent deep reinforcement learning has been applied to address a variety of complex problems with either discrete or continuous action spaces and achieved great success. However, most real-world environments cannot be described by only discrete action spaces or only continuous action spaces. And there are few works having ever utilized deep reinforcement learning (drl) to multi-agent problems with hybrid action spaces. Therefore, we propose a novel algorithm: Deep Multi-Agent Hybrid Soft Actor-Critic (MAHSAC) to fill this gap. This algorithm follows the centralized training but decentralized execution (CTDE) paradigm, and extend the Soft Actor-Critic algorithm (SAC) to handle hybrid action space problems in Multi-Agent environments based on maximum entropy. Our experiences are running on an easy multi-agent particle world with a continuous observation and discrete action space, along with some basic simulated physics. The experimental results show that MAHSAC has good performance in training speed, stability, and anti-interference ability. At the same time, it outperforms existing independent deep hybrid learning method in cooperative scenarios and competitive scenarios.}

\keywords{multi-agent, deep reinforcement learning, hybrid action spaces, maximum entropy}



\maketitle

\section{Introduction}\label{sec1}

In recent years, deep reinforcement learning (DRL) algorithm\cite{bib1,bib2} has been applied in many multi-agent fields to handle practical tasks, such as multiplayer games, autonomous driving, and the research of it has made great progress. This is crucial to building artificially intelligent systems that can effectively interact with humans and each other.

However, there are many problems and challenges about multi-agent reinforcement learning (MADRL)\cite{bib3,bib4}. Contrary to the theory, the realistic environment is changeable. In many settings, owing to restricted communication and partial observability, one agent can only realize the environmental changes, but ignore the affection of other agents' actions on its own actions. At the same time, the communications between agents have a great impact on theirselves to learn an appropriate method. Fortunately, we can train our agents using the paradigm of centralized training and decentralized execution which is common in a variety of popular MADRL algorithms to get good policies in such environments. The well-known MADDPG\cite{bib5} which adopts the qmix network\cite{bib6} to estimates the joint action value as a complex nonlinear combination of the action values of each agent only based on local observations is the representative of this idea.

On the other hand, most popular multi-agent reinforcement learning algorithms ask the action spaces to be discrete or continuous which is not consistent with the real world that the action space is usually discrete-continuous hybrid, such as Real Time Strategic (RTS) games and Robot movements. In this environment, each agent needs to select a discrete operation and its related continuous parameters in each time step. In order to solve this problem, the common practice is to simply approximate the mixed action space with a discrete set, or relax it into a continuous set which has a bad effection for many limitations. For establishing a good approximation of the continuous part of hybrid action, a huge number of discrete actions which may significantly increase the complexity of the action space are often necessary.

Then, a better solution is to learn directly in hybrid action spaces\cite{bib7,bib8}. Hybrid Soft Actor-Critic (HSAC) is a successful algorithm to solve hybrid action space problem following this mind. However, the attempt to apply it to multi-agent setting is not ideal because of nonstationarity in multi-agent environments.

In this work, we propose a novel approach to address multi-agent problems in discrete-continuous hybrid action spaces with centralized training and decentralized execution framework: Deep Multi-Agent Hybrid Soft Actor-Critic (MAHSAC) algorithm. We extend HSAC to multi-agent settings, the policy network of each agent outputs random actions represented by Gaussian distribution. Our approach has strong exploration capability and generalization, due to we explored the best possibilities in different ways. So, it is easier to adjust when encountering interference, and the learned strategies can be used as initialization for more complex and specific tasks. Empirical results on Multi-Agent Particle Environment which is a simple multi-agent particle world show the superior performance of our approach compared to decentralized HSAC in both cooperative and competitive environment.

\section{Background and related work}\label{sec2}

\subsection{Deep Multi-agent Reinforcement Learning}\label{subsec2}

\begin{figure}[h]%
\centering
\includegraphics[width=0.9\textwidth]{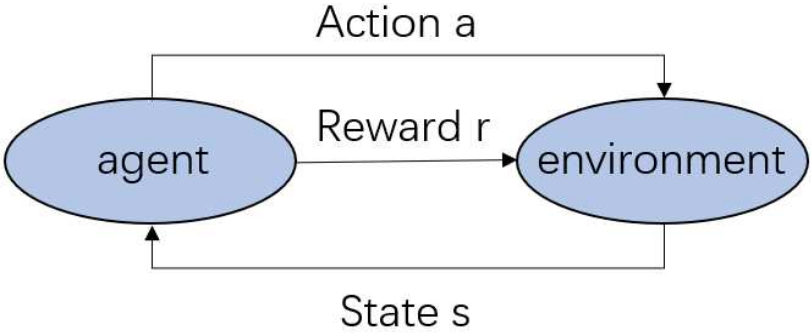}
\caption{the framework of deep reinforcement learning}\label{fig1}
\end{figure}

Drl has the abilities of feature extraction and sequence decision making which can solve many complex decision problems. It often uses Markov decision models to decompose problems and collects the state $s_t$, action $a_t$, reward $r_t$, the next state $s_{t+1}$ as a tuple ($s_t$ , $a_t$ , $r_t$ , $s_{t+1}$) at current time step to form a set ($S$,$A$,$R$,$S'$). The structure is shown in Figure \ref{fig1}. The optimization objective is strategy $\pi: s \rightarrow {a}$, and the cumulative reward which obtained at the time t by formula \ref{eq1} is maximized by optimizing it.

\begin{equation}
R=\sum\limits_{t'=t}^{T}y^{t'-t}r_t \label{eq1}
\end{equation}

Where $\gamma$ represents the attenuation factor.

Then, the Q function can be defined as $Q^\pi=E\left[R_t\mid s_t,a_t \right]$, and the optimal strategy $\pi^*$ is selected as the optimization goal to maximize the expectation of formula \ref{eq1}, which means that the formula \ref{eq2} is accurate.

\begin{equation}
Q^{\pi^*}\left(s,a\right)\geq Q^\pi\left(s,a\right) \forall s,a \in S,A \label{eq2}
\end{equation}

When deep reinforcement learning is utilized to multi-agent systems, the centralized training and decentralized execution is usually the most popular framework at both discrete action and continuous action spaces, where MADDPG and Qmix are representative ones.

MADDPG focuses on multi-agent problems with continuous action spaces which is aimed at learning a centralized critic
to take as input the actions of all agents and global state information. Qmix focuses on multi-agent problems with discrete action spaces which employs a mixing network to estimate joint action-values as a complex non-linear combination of per-agent action value that conditions only on local observations. The weights of the mixing network are produced by separate hypernetworks which take the global state information as input.

However, most of the existing multi-agent algorithms focus on only discrete action space or continuous action space which cannot be applied to many real-world environments. More effective algorithms for multi-agent problems with hybrid action space are still needed.

\subsection{Hybrid Soft Actor-Critic}\label{subsec3}

SAC\cite{bib9,bib10} is an exceptional off-policy algorithm that was originally proposed for continuous action space.
This method aims to maximize both return of the agents and entropy of the actions which is significantly different from traditional algorithms which pay attention to maximize return of the agents. The main idea is to add an entropy bonus to the objective optimized by the agent, i.e. maximizing

\begin{equation}
E_\pi \left[\sum\nolimits_t \gamma^t \left(r_t+\alpha\mathcal{H}\left(\pi\left(\cdot\mid s_t\right)\right) \right)\right] \label{eq3}
\end{equation}

which can prevent the agent's strategy from converging prematurely to the local optimal solution. Where, policy $\pi$ is used to calculate distributions of the trajectory ($s_t,a_t$) from the strategy, that is, the entropy $\mathcal{H}$($\cdot$) of the strategy. $\alpha$ is used to pay more attention to entropy or reward, and a higher $\alpha$ will especially be conducive to explore possibility by encouraging agents to take actions which are more random. According to the soft policy iteration method, SAC updates the participant network by minimizing KL differences, and the following objectives need to be minimized:

\begin{equation}
J_\pi \left(\theta\right)=E_{s_t\sim D,a_t\sim\pi_\theta}\left[\alpha log\left(\pi_\theta\left(a_t\mid s_t\right)\right)-Q_\beta \left(s_t,a_t\right) \right]\label{eq4}
\end{equation}

Where, $\beta$ is the parameter of the critic network $Q_\beta$, $\theta$ is the parameter of the agent actor network $\pi_\theta$, and D represents the replay buffer. And it is failure to update the process by back propagating parameters because $a_t$ used in $Q_\beta(s_t,a_t)$ is obtained by sampling random policies $\pi_\theta$. Therefore, the reparameterization technique is practical for minimizing $J_\pi(\theta)$ by the random gradient. By adding random noise $\xi$ which follows the Gauss distribution, we can sample with a Gauss distribution. And then, the original sampling process is changed to: 

\begin{equation}
\Tilde{a}_\theta \left(s_t,\xi \right)=\tanh\left(\mu_\theta\left(s_t\right)+\sigma_\theta \left(s_t\right)\odot\xi \right)\label{eq5}
\end{equation}

Where, $\Tilde{a}_\theta \left(s_t,\xi \right)$ is the sample of probability distribution of agent's behavior $\pi_\theta\left(\cdot\mid s_t\right)$ , $\mu_\theta$ is mean value and $\sigma_\theta$ represents the standard deviation. Besides, $\tanh$ is used to ensure that the value of the agent's operation is limited within a certain range. By this approach, the optimization objectives are changed to:

\begin{equation}
\nabla_\theta J_\pi\left(\theta\right)=\nabla_\theta\alpha \log\left(\pi_\theta\left(a_t\mid s_t\right)\right)+\left(\nabla_{a_t}\alpha \log\left(\pi_\theta\left(a_t\mid s_t\right)\right)-\nabla_{a_t}Q\left(s_t,a_t\right)\right)\nabla_\theta\Tilde{a}_\theta \left(s_t,\xi \right) \label{eq6}
\end{equation}

And the critic network can be updated by minimizing the Berman error:

\begin{align}
J_Q\left(\beta\right)&=E_{s_t,a_t,r_t,s_{t+1}\sim D}\left[\frac{1}{2}\left(Q_\beta\left(s_t,a_t\right)-y\right)^2\right]\label{eq7}\\
y&=r\left(s_t,a_t\right)+\gamma E_{a_{t+1}\sim\pi_\theta\left(s_{t+1}\right)}\left[Q_{\overline{\beta}}\left(s_{t+1},a_{t+1}\right)-\alpha\log\left(\pi_\theta \left(a_{t+1}\mid s_{t+1}\right)\right)\right] \label{eq8}
\end{align}

To deal with the problem of mixing discrete and continuous actions in SAC, there is a setting of representing agent's operation a by a combination of discrete components $a^d= (a^d_1, . . . , a^d_D)$ and continuous components $a^c= (a^c_1, . . . , a^c_C)$. Each $a^d_i$ is an integer which represents the i-th discrete action that the agent may take. Each $a^c_j$ is a continuous vector which represents its j-th continuous action\cite{bib11,bib12}. When the observed state s is given, assuming the discrete components are independent. And the continuous components are independent while both s and the discrete actions are given. Then, yielding the following decomposition:

\begin{equation}
\pi\left(a\mid s\right)=\pi\left(a^d \mid s\right)\pi\left(a^c \mid s,a^d \right)
\label{eq9}
\end{equation}

Here, the same letter $\pi$ is used to represent both discrete probability mass functions and probability density functions which can be applicable in different components of actions. According to the differences of the number of discrete actions and continuous actions as well as their relationship, there are many different action representations.

\begin{figure}[h]%
\centering
\includegraphics[width=0.9\textwidth]{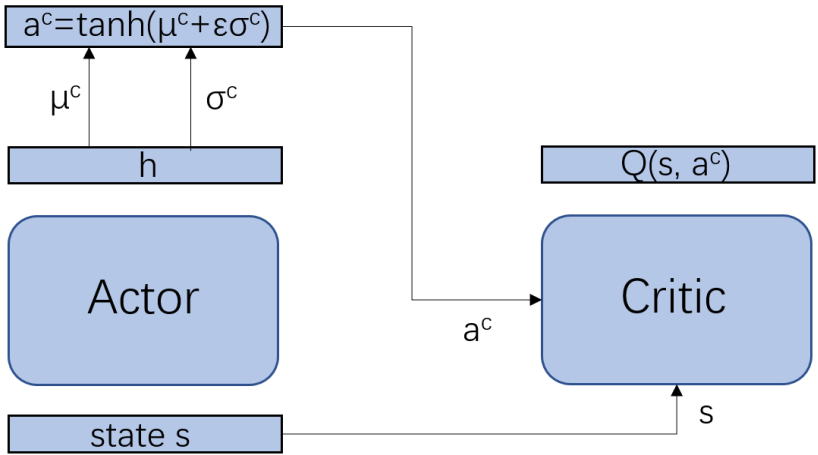}
\caption{architecture of standard SAC}\label{fig2}
\end{figure}

Figure \ref{fig2} shows the typical architecture of standard continuous SAC. By injecting standard normal noise $\xi$ and applying $\tanh$ nonlinearity to keep the action within a bounded range, actors can output the mean and standard deviation vectors $\mu^c$ and $\sigma^c$ which are used to sample action $a^c$. Then, the critic can estimate the corresponding Q-value by taking both state s and actor’s action $a^c$. 
 
\begin{figure}[h]%
\centering
\includegraphics[width=0.9\textwidth]{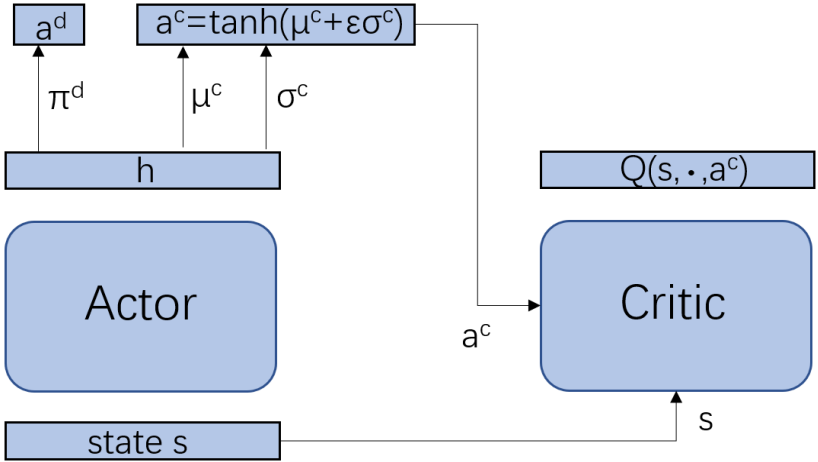}
\caption{architecture of HSAC}\label{fig3}
\end{figure}

SAC with hybrid action space requires a different policy parameterization which calls for a different network architecture. Therefore, Figure \ref{fig3} shows a situation where the agent must combine a discrete action $a^d$ with a set of independently sampled continuous parameters $a^c$. Here, a shared hidden state representation h produces additionally a discrete distribution $\pi^d$ to sample the discrete action $a^d$.
Note that here, the value from critic’s output layer contains all discrete actions' predicted Q-values.
 
The SAC algorithm is based on the idea that the entropy additive value proportional to the entropy of $\pi(a\mid s)$ is given. As long as the action has a discrete part, the joint entropy definition with the weighted sum of discrete and continuous actions becomes:

\begin{equation}
\mathcal{H}(\pi(a^d,a^c\mid s))=\alpha^d \mathcal{H}(\pi(a^d\mid s))+\alpha^c \sum_{a^d}\pi(a^d\mid s)\mathcal{H}(\pi(a^c\mid a^d, s))
\label{eq10}
\end{equation}

Among them, the hyperparameters $\alpha^d$ and $\alpha^c$ encourage the exploration of discrete and continuous actions respectively. Besides, these two hyperparameters can be automatically adjusted in the learning process by using the same optimization techniques as soft actor-critic algorithms. Otherwise, they can be set to the same fixed value.

\section{Method}\label{sec3}

To deal with hybrid action spaces problems in multi-agent environment, one attempt is to equip each agent with a decentralized HSAC algorithm in the independent learning paradigm which will be seriously affected by the the non-stationarity of environments and gets not good results in practice. So, in this paper, we propose a deep multi-agent learning method--Deep MAHSAC for hybrid action spaces which extends the HSAC and learns effective coordination strategies directly in the hybrid action spaces.

The main factor for the failure of the attempt to extend directly the single-agent deep reinforcement learning algorithm to the multi-agent system is that each agent has a close relationship with other agents and the policy of them is ever-changing. Therefore, the environment is unstable or does not conform to Markov properties. To solve this problem, just as MADDPG extends DDPG into the multi-agent system, we employ the CTDE framework which introduce a global network of critics who direct the training of distributed actor networks during running, while execution performed by actors using only local observations to take actions.

We extend HSAC to multi-agent setting based on CTDE training paradigm, and the network model as shown in figure \ref{fig4}. We set up N agents in the environment to suit various situations, and the actor networks of these agents are a set: $\pi=\{\pi_1,\pi_2,...,\pi_n\}$, while the $Q^\beta_i$ represents the parameters of critic network $\beta_i$. For each agent $\pi_i$, it gets local observation $o_i$ from the environment as input and output the action $a_i$. In this paradigm, we can update the actor network of agent i by minimizing the target:

\begin{figure}[h]%
\centering
\includegraphics[width=0.9\textwidth]{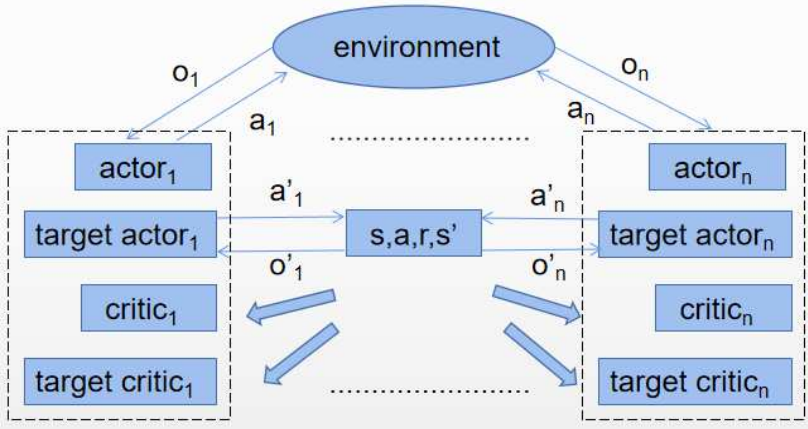}
\caption{The architecture of MAHSAC}\label{fig4}
\end{figure}

\begin{equation}
J_{\pi_i}(\theta_i)=E_{s\sim D,a\sim\pi_\theta}[\mathcal{H}(\pi(a_i\mid o_i))-Q_\beta (s,a)]
\label{eq11}
\end{equation}

Where, the joint entropy $\mathcal{H}(\pi(a_i\mid o_i))$ is defined as the formula \ref{eq10}. The state $s=\{o_1,o_2,...,o_n\}$ which is a set of agent i‘s local observation $o_i$ and the action $a=\{a_1,a_2,...,a_n\}$ which is a set of agent i‘s hybrid action $a_i$ are both stored in the experience replay buffer D. Each agent i has its own independent global critic network which takes the combination set of states and actions from all agents as input, and the actor network takes the Q value corresponding to actions taken by agent i at timestep t as output to train strategy $\pi_i$ during training. Especially, action a taken as input by $Q_\beta$ is a set of continuous components $a_c$ from all agents. After the training, each agent only needs to observe its own local observation value $o_i$ to get a random strategy represented by the Gauss distribution. Besides, we can update agent i’s global critic network by minimizing the following:

\begin{equation}
J_Q (\beta_i)=E_{s,a,r,s'}[\frac{1}{2}(Q^{\beta_i}(s,a)-y)^2]
\label{eq12}
\end{equation}

Where, y is represented as:

\begin{equation}
y=r_i+\gamma E_{a'\sim \pi_{\overline{\theta}_i}}[Q^{\overline{\beta}_i}(s',a')-\mathcal{H}(\pi(a'_i\mid o'_i))]
\label{eq13}
\end{equation}

Here, $\gamma$ is the discount factor.

Our algorithm makes use of two soft Q-functions to mitigate positive bias in the policy improvement step that will degrade performance of value-based methods. We parameterize two soft Q-functions with parameters $\theta_i$, and calculate the Bellman error $J_{Q_i}$ of them separately. And the final $J_Q$ is just their average value. By this way, we can speed up training, especially on harder or complex tasks.

In order to stabilize the training of deep neural network, each agent needs to add the actor target network $\pi_{\overline{\theta}_i}$ and critic target network $Q_{\overline{\beta}_i}$. Those target network parameters will be updated in soft mode every few episodes:

\begin{equation}
\overline{\theta}_i=\tau\theta_i+(1-\tau)\overline{\theta}_i\\
\label{eq14}
\end{equation}
\begin{equation}
\overline{\beta}_i=\tau\beta_i+(1-\tau)\overline{\beta}_i
\label{eq15}
\end{equation}

Obviously, the hyper-parameter $\tau$ can significantly affect the update of target network parameters.

\section{Experiment}\label{sec4}

\subsection{Experimental Environment}\label{subsec4}

In our experiment, we adopt a simple multi-agent particle environment which consists of N agents and L landmarks inhabiting a two-dimensional world with continuous space and discrete time. We have modified this environment so that it can meet the demand of continuous-discrete hybrid action space. Agents may take physical actions and communication actions that get broadcasted to other agents in the environment. There are some different scenarios with two different topics in the environment and MHASAC was evaluated using decentralized HSAC as a baseline in the two topics: cooperative and competitive. 

\begin{figure}[h]%
\centering
\includegraphics[width=0.5\textwidth]{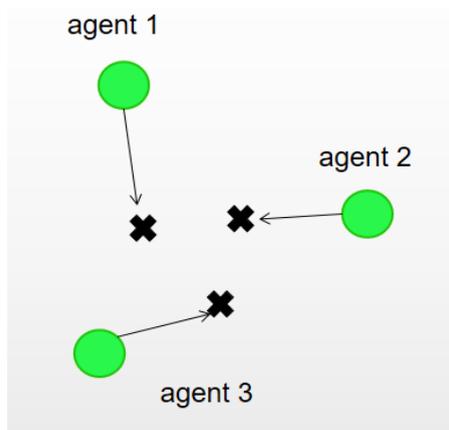}
\caption{Cooperative Navigation Scenario}\label{fig5}
\end{figure}

As shown in the figure \ref{fig5}, there are three agents and three target points in a two-dimensional plane for the cooperative navigation scenario. Agents observe the relative positions of other agents and target points. Then, they will be collectively rewarded based on the proximity of any agent to each target. In addition, the agents occupy significant physical space and will be punished when they collide with each other. These agents need to learn to infer which targets they must cover and move there while avoiding other agents.

\begin{figure}[h]%
\centering
\includegraphics[width=0.5\textwidth]{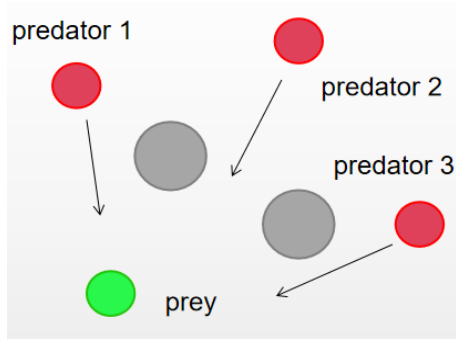}
\caption{Predator-Prey Scenario}\label{fig6}
\end{figure}

As we can see in figure \ref{fig6}, there are 3 predators, 1 prey, and 2 obstacles in a two-dimensional plane for the Predator-Prey scenario. The goal of predators is to cooperate with each other to capture the prey, while the prey is aimed at avoiding the prey. Therefore, there is competition between the predators and the prey, and there is cooperation between the predators. 

\subsection{Experimental Result}\label{subsec5}

In the cooperative navigation scenario, agents will be rewarded according to the distance from the target and be punished when they collide with each other. Therefore, in this scenario, MAHSAC and decentralized HSAC training models were compared in terms of the reward, number of collisions and distance from the target.

\begin{figure}[h]%
\centering
\includegraphics[width=0.9\textwidth]{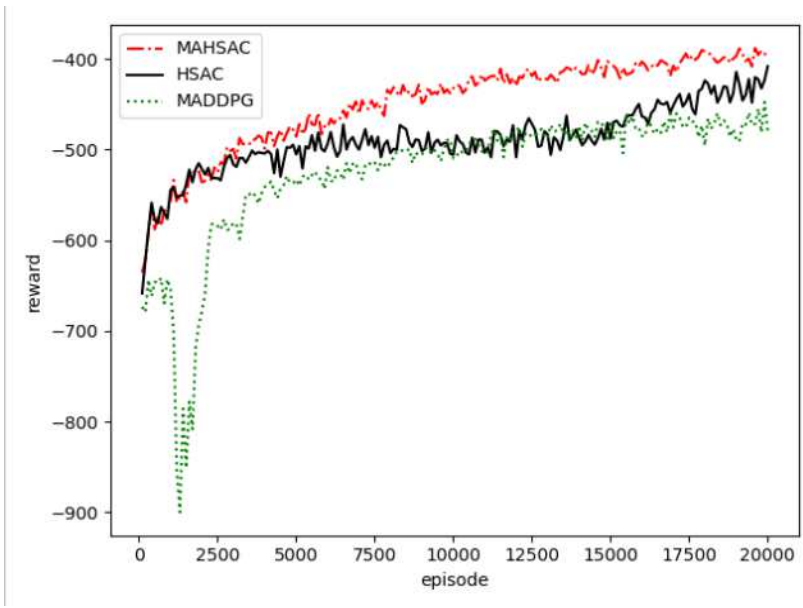}
\caption{agent reward on cooperative communication after 20000 episodes}\label{fig7}
\end{figure}

After 20000 episodes of training, the reward value curve of the agents is shown in figure \ref{fig7}. We recorded the reward of the sum of all 3 agents per episode and save the average value every hundred episodes. Besides, we also applied MADDPG model to train in the original multi-agent particle environment as a comparison. As the fig shows, MAHSAC significantly has a better performance than decentralized HSAC and MADDPG in training speed, reward and stability.

Table \ref{tab1} shows the number of collisions and distance from the target after algorithm convergence, which corresponds to the performance test of agents. Compared with decentralized HSAC, MAHSAC has lower values in these two respects which means its better performance.

\begin{table}[h]
\begin{center}
\begin{minipage}{174pt}
\caption{ The average number of conflicts and the average proxy distance from landmarks per episode in the
cooperative navigation scenario}\label{tab1}%
\begin{tabular}{@{}llll@{}}
\toprule
agent & collisions  & dist \\
\midrule
MAHSAC    & 1.84395   & 0.242679  \\
HSAC    & 2.085   & 0.323587  \\
\botrule
\end{tabular}
\end{minipage}
\end{center}
\end{table}

The above data shows that MAHSAC can train policy that will achieve the goal faster and better in the cooperative environment than decentralized HSAC.

And in the Predator-Prey scenario, due to the competition between predators and prey, the reward value curve of the agents is unstable and cannot reflect the real performance of algorithms. Therefore, we recorded the average number of prey touches by predator per episode. Here, in order to concretely show the difference in the two algorithms' performance, we set four situations where MAHSAC and decentralized HSAC were respectively applied to predator and prey agents.

 The test results are shown as table \ref{tab2}, a higher number of collisions indicates that the predators can catch their prey earlier. As we can see, MAHSAC predators are far more successful at chasing HSAC prey than the converse and HSAC had a bad performance when directly pitted against MAHSAC in all cases.

\begin{table}[h]
\begin{center}
\begin{minipage}{174pt}
\caption{ Average number of prey touches by predator per episode on the two predator-prey scenario}\label{tab2}%
\begin{tabular}{@{}llll@{}}
\toprule
agent & adversary & touches  \\
\midrule
MAHSAC    & MAHSAC   & 2.89785  \\
MAHSAC    & HSAC   & 20.587596  \\
HSAC    & MAHSAC   & 2.2201  \\
HSAC    & HSAC   & 1.8726  \\
\botrule
\end{tabular}
\end{minipage}
\end{center}
\end{table}

\section{Conclusion}\label{sec5}

This paper makes an attempt to extend HSAC to multi-agent settings and provides a novel way to apply deep reinforcement learning in multi-agent environments to handle practical problems with discrete-continuous hybrid action spaces which further fills the vacancy in this area. Under the paradigm of centralized training and decentralized execution, we propose MAHSAC, and the experimental results show its superiority to decentralized HSAC method under an easy multi-agent particle world basic simulated physics with cooperative and competitive scenarios. For future work, we wish to extend our algorithm to the scenario which agent can input and receive communication output to each other. By this way, our algorithm can adapt to more realistic scenes and make more good performance. In addition, further improving the training speed and stability of the algorithm is important.

\end{document}